\documentclass[journal]{IEEEtran}
\ifCLASSINFOpdf
  \usepackage{graphicx}
  \graphicspath{{images/}}
\else
\fi
\usepackage{placeins}
\usepackage{amsmath}
\usepackage{amssymb}
\usepackage{algorithmic}
\usepackage{algorithm}
\usepackage{booktabs}
\usepackage{threeparttable}
\usepackage{multirow}
\newcounter{savedfigure}

\begin{document}
%
\title{SAM2Dual: Training-Free, Dual Memory for Long-Term Video Object Segmentation}
%
%
%

\author{JeongRae~Kim and Changwon~Lim\textsuperscript{*}%
\thanks{J.~Kim is the first author and C.~Lim\textsuperscript{*} is the corresponding author.}%
\thanks{Both authors are with the Department of Applied Statistics, Chung-Ang University, Seoul 06974, Republic of Korea.}%
}

%
%

\markboth{IEEE TRANSACTIONS ON IMAGE PROCESSING}%
{SAM2Dual}%

%



\maketitle

\begin{abstract}
Long-term video object segmentation (VOS) remains challenging due to error accumulation under
extended occlusions, re-appearance, and scene changes.
Although SAM2 provides strong zero-shot performance, its streaming memory can amplify drift over
long horizons when recent, unreliable predictions dominate the memory state.
We propose \textbf{SAM2Dual}, a \textbf{training-free, plug-and-play} inference-time enhancement
that improves long-video robustness \textbf{without updating model weights}.
SAM2Dual introduces a \textbf{Dual Memory} design that explicitly separates
(i) \emph{short-term} memory for rapid local adaptation and
(ii) \emph{long-term} memory built via interval-based sampling to preserve global identity cues,
combined through a gated fusion strategy.
In addition, we present \textbf{Text-Aware Memory (TAM)}, which extracts a compact word-level cue
from early frames and uses text embeddings to reweight memory contributions based on semantic
compatibility, supporting identity preservation when visual evidence becomes weak or ambiguous.
Across long-term benchmarks, SAM2Dual consistently improves stability on long videos, raising
J\&F from 49.33 to 50.65 on MOSEv2 and achieving consistent gains on LVOSv2.
\end{abstract}


%
\IEEEpeerreviewmaketitle

\section{Introduction}
%
%
%
%
\IEEEPARstart{V}{ideo} Object Segmentation (VOS) aims to produce pixel-accurate masks for target objects over time.
Recent progress has been driven by two complementary directions:
memory-based propagation methods that store and retrieve object information across frames,
and foundation models that offer strong zero-shot generalization through large-scale pretraining.
While early benchmarks and methods largely focused on short clips, recent datasets and real-world scenarios increasingly involve long, unconstrained videos.
In such settings, targets may undergo prolonged occlusions, abrupt changes in pose or illumination, motion blur, camera viewpoint shifts, and even scene transitions, making errors prone to accumulate over time.
Consequently, robust VOS is not simply a matter of strong per-frame segmentation; it requires maintaining consistent instance identity and recovering from long-horizon drift and uncertainty.

\begin{figure*}[t]
  \centering
  \includegraphics[width=\textwidth]{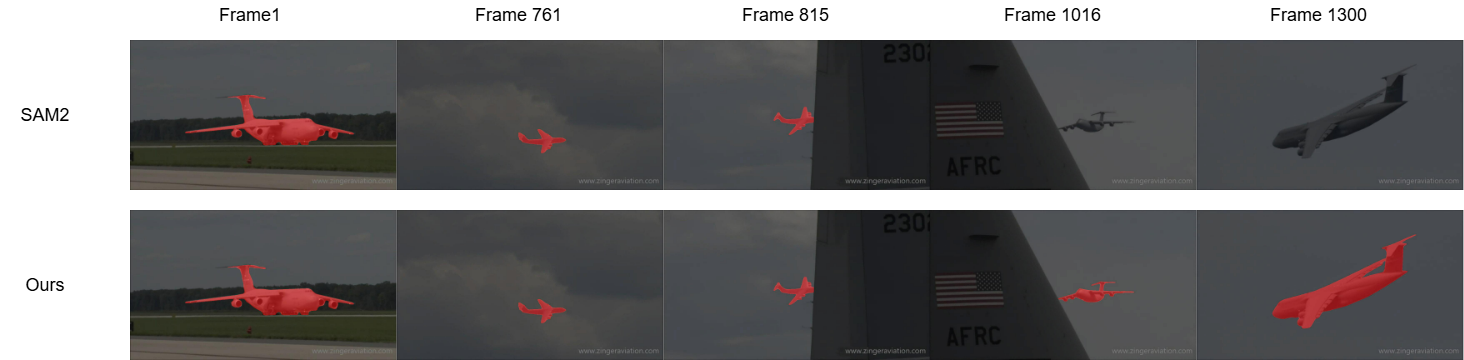}
  \caption{A qualitative comparison on a long sequence (over 1,300 frames) shows that, under the same first-frame mask initialization, SAM2 progressively drifts and loses the target in later frames (e.g., Frames 1016 and 1300) under large scale and pose variations, whereas the proposed approach maintains consistent segmentation throughout the video, effectively suppressing long-term drift.}
  \label{fig:plane}
\end{figure*}

A central challenge in long-term VOS is \emph{error accumulation}.
Many state-of-the-art systems operate iteratively, where predictions at time $t$ serve as evidence for time $t{+}1$.
Under partial occlusion or visual ambiguity, the tracker may drift to distractor regions or cause an identity switch.
Once an incorrect mask or corrupted feature is written into memory, subsequent propagation often amplifies the error, making recovery progressively harder.
This issue is particularly severe in long sequences, where the target can disappear and reappear after many frames, or where substantial appearance changes render earlier cues unreliable.
Therefore, long-term robustness calls for explicit mechanisms to mitigate memory contamination, prevent unreliable evidence from dominating future inference, and preserve identity cues across large temporal gaps.
An example is shown in Fig.~\ref{fig:plane}, where SAM2 gradually loses the target in a long sequence with large scale and pose variations~\cite{Sam2}.

Memory-based VOS frameworks attempt to handle temporal variation
by storing features or masks from previous frames and matching them to the current frame.
However, practical memory designs face an inherent trade-off:
recent frames provide strong local evidence for short-term adaptation
but are vulnerable to transient noise such as occlusion,
while older frames preserve global identity information
but can become mismatched when the object appearance changes significantly.
Systems that rely on a single streaming memory are particularly sensitive to this tension~\cite{Xmem}.
If the memory is dominated by recent observations, the model may overfit to short-lived cues and drift;
if the memory is overly conservative, the model may fail to adapt to legitimate appearance changes.
This suggests that a monolithic memory update is an incomplete solution for long videos,
and that temporal evidence should be structured explicitly
to balance fast adaptation and long-range stability.

Recent Segment-Anything-style video models (e.g., SAM2-style architectures)
have demonstrated impressive generalization and can be deployed without per-dataset training.
Nevertheless, strong zero-shot segmentation does not automatically translate into stable long-term tracking.
In challenging videos, the model must repeatedly answer not only ``where is the object now?''
but also ``which object is it?''
when multiple similar instances exist,
when the target reappears after a long absence,
or when only partial visual evidence is available.
Purely visual matching can be insufficient in such cases,
motivating the use of lightweight semantic cues as an identity anchor.
Even minimal word-level semantics can provide an additional constraint
that complements appearance-based evidence,
especially when the visual signal is weak, intermittent, or confusable.

In this work, we propose \textbf{SAM2Dual},
a \textbf{training-free, plug-and-play} inference-time framework
that improves robustness on long videos \textbf{without updating model weights}.
Our approach follows two principles:
explicitly structure temporal evidence to mitigate the short-term/long-term trade-off in memory,
and inject lightweight semantics to support identity preservation under uncertainty.
First, we introduce a \textbf{Dual Memory} design that separates
(i) a \textbf{short-term} memory emphasizing recent frames for rapid local adaptation and
(ii) a \textbf{long-term} memory that stores interval-sampled frames
to preserve global identity cues over long horizons.
A gating mechanism dynamically balances these memories,
reducing drift while retaining flexibility under appearance changes.
Second, we present a \textbf{Text-Aware Memory (TAM)} module
that extracts compact word-level cues from early frames
and uses text embeddings to reweight memory contributions based on semantic compatibility,
reinforcing identity consistency when appearance cues degrade.
We evaluate SAM2Dual on a diverse suite of short- and long-term VOS benchmarks
and observe consistent improvements in long-video stability
while maintaining competitive performance on standard short-sequence datasets.

 

\section{Related Works}

\subsection{Foundation models for promptable video segmentation.}
Segment Anything Model~2 (SAM2) extends promptable segmentation from images to videos
with a streaming-memory mechanism, enabling strong zero-shot performance
and practical inference-time deployment~\cite{Sam2}.
By formulating video segmentation as a prompt-driven process,
SAM2 serves as a strong backbone for interactive or semi-automatic VOS pipelines,
where limited user input can be propagated across frames.
However, long and unconstrained videos remain challenging:
temporal uncertainty (e.g., prolonged occlusion, re-appearance, and scene changes)
can induce drift and amplify errors through iterative propagation.
In particular, when recent predictions become unreliable, streaming updates may over-emphasize short-lived cues,
making recovery increasingly difficult over long horizons.
This motivates inference-time strategies that improve long-horizon robustness
without additional training or weight updates.

\subsection{Long-term memory design in video object segmentation.}
A dominant line of VOS research improves temporal consistency
by storing and retrieving object information from previous frames.
Prior works explore various memory management strategies to balance robustness and efficiency in long videos.
XMem proposes a long-video VOS framework inspired by the Atkinson--Shiffrin memory model,
introducing interacting memory stores and a consolidation mechanism
that transfers useful information into long-term memory~\cite{Xmem}.
This design aims to reduce long-horizon performance decay
while controlling memory growth over extended sequences.
XMem further highlights that explicitly separating short-lived evidence
from sustained identity cues is beneficial for long-term robustness.
DeAOT builds on AOT-style association transformers for efficient propagation~\cite{AOT,DEAOT}.
Despite these advances, many long-term memory designs rely on dedicated training or architectural changes,
and their integration into foundation-style promptable backbones is not always straightforward.
Our work follows this principle, but targets a different setting:
we propose a \emph{training-free} enhancement that can be plugged into a SAM2-style backbone
at inference time, improving stability without redesigning or retraining the base model.

\subsection{Vision--language representations for semantic anchoring.}
CLIP learns aligned image--text embeddings from large-scale paired data
and enables zero-shot transfer by comparing visual features with textual concepts~\cite{CLIP}.
Such vision--language representations provide a lightweight semantic signal
that complements appearance-based matching when visual evidence becomes weak or ambiguous.
In VOS, even a compact word-level cue can act as an identity anchor,
reducing confusion among similar instances and supporting recovery after long occlusions.
In our setting, we adopt JINA-CLIP~\cite{JINA-CLIP}, whose Matryoshka-style training~\cite{matryoshka} encourages compact embeddings to retain semantic information,
making it suitable for training-free inference-time modulation with minimal information loss.
However, purely visual matching can still be insufficient when early-frame cues are weak or the target is fine-grained,
motivating auxiliary semantic constraints that stabilize identity over time.
Motivated by this, we leverage CLIP-style text embeddings
not as a primary segmentation driver,
but as an auxiliary signal that modulates memory usage
to encourage identity-consistent propagation under uncertainty.

\section{Method}
\label{sec:method}

\begin{figure*}[t]
  \centering
  \includegraphics[width=\textwidth]{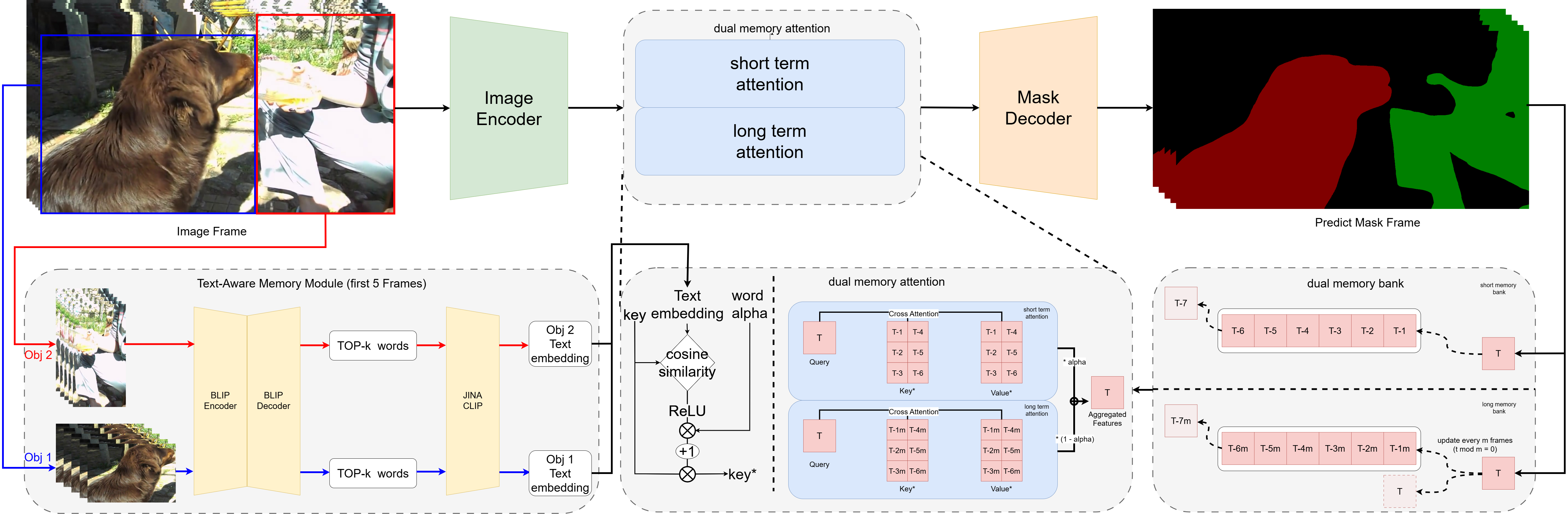}
  \caption{Overall architecture of the proposed method. Given video frames, the model reads from short-term (ST) and long-term (LT) memories, fuses the readouts, and decodes masks for each frame.
  Notation: $m$: long-term update interval; $\alpha$: ST/LT fusion gate;
  $\lambda_{\mathrm{word}}$: text modulation strength; $g$: text gate.}
  \label{fig:architecture}
\end{figure*}

\subsection{Problem Setup and Overview}
Given a video sequence $\{x_t\}_{t=1}^{T}$,
we assume a first-frame prompt mask $y_1$ is provided for the target object.
Our goal is to predict a mask sequence $\{\hat{y}_t\}_{t=2}^{T}$
in a streaming (online) manner without any additional training.
Following the SAM2 inference paradigm~\cite{Sam2},
the first-frame prompt is converted into a memory slot $m_1$ via a memory encoder,
and each subsequent frame is processed by attending to stored memory tokens
to update the segmentation prediction.

For multi-object datasets, we apply SAM2Dual in an object-wise manner:
each object $o$ maintains its own memory banks $(M^{ST}_{t,o}, M^{LT}_{t,o})$ and word cue $w_o^\ast$.
Final scores are obtained by averaging over objects following the standard J\&F protocol~\cite{DAVIS_Ori}.

We propose \textbf{SAM2Dual},
an inference-time plug-in framework built on top of SAM2~\cite{Sam2},
composed of two complementary modules:
(1) \textbf{Dual Memory}, which restructures the baseline single streaming memory
into short-term (ST) and long-term (LT) banks to mitigate long-range drift;
and (2) \textbf{Text-Aware Memory (TAM)},
which injects a lightweight semantic identity cue
to modulate memory usage when visual evidence is weak or ambiguous.
The overall pipeline is training-free
and preserves the original SAM2 backbone and decoder.

\subsection{Dual Memory}
\label{sec:dual_memory}
The key idea of Dual Memory is to separate temporal evidence
into two different time scales:
a short-term memory that captures rapid local appearance changes,
and a long-term memory that preserves global identity cues across large temporal gaps.
Let $m_{t,o}$ denote the memory slot constructed from frame $t$ for object $o$
(e.g., the key/value tokens produced by the SAM2 memory encoder).
At time $t$, we maintain two memory banks:

\begin{equation}
\begin{aligned}
M_{t,o}^{ST} &= \{m_{t-K,o}, \dots, m_{t-1,o}\}, \\
M_{t,o}^{LT} &= \{m_{t-m,o}, m_{t-2m,o}, \dots, m_{t-Lm,o}\}.
\end{aligned}
\label{eq:dual_memory_banks}
\end{equation}

Here, $K$ is the short-term memory length,
$m$ is the long-term sampling (update) interval,
and $L$ is the maximum number of long-term slots.
Importantly, both ST and LT banks are queried using the same (frozen) SAM2 memory attention module;
they share identical weights and differ only in the temporal composition of stored slots.
This design enables two temporal contexts without any additional training:
all SAM2 parameters remain frozen, and we only reorganize the streaming memory into ST/LT banks at inference time.

\subsubsection{Short-Term Memory Read}
The short-term (ST) bank stores the most recent $K$ slots.
Given the query feature $q_{t,o}$ extracted from the current frame $x_t$ for object $o$,
we obtain an ST memory response via memory attention:
\begin{equation}
\begin{aligned}
Z^{ST}_{t,o} = \mathrm{Attn}(q_{t,o}, M^{ST}_{t,o}).
\end{aligned}
\label{eq:st_read}
\end{equation}
This corresponds to explicitly defining the baseline ``recent-frame'' memory usage
as a dedicated short-term bank.

\subsubsection{Long-Term Memory Update and Read}
The long-term (LT) bank stores temporally distant slots at a fixed interval.
To control memory growth and preserve global context,
the LT bank is updated only when $t$ is a multiple of $m$:
\begin{equation}
\begin{aligned}
M^{\mathrm{LT}}_{t+1,o} =
\left\{
\begin{array}{ll}
\mathrm{Update}\!\left(M^{\mathrm{LT}}_{t,o}, m_{t,o}\right), & t \bmod m = 0,\\[2pt]
M^{\mathrm{LT}}_{t,o}, & \mbox{otherwise}.
\end{array}
\right.
\end{aligned}
\label{eq:lt_update}
\end{equation}
The $\mathrm{Update}(\cdot)$ operator appends $m_{t,o}$ to the LT bank
and retains at most $L$ slots (e.g., oldest-first eviction).
Using the same query $q_{t,o}$,
we read the LT response as:
\begin{equation}
\begin{aligned}
Z^{LT}_{t,o} = \mathrm{Attn}(q_{t,o}, M^{LT}_{t,o}).
\end{aligned}
\label{eq:lt_read}
\end{equation}
Intuitively, $Z^{LT}_{t,o}$ provides a stable identity reference
from temporally distant frames,
which becomes particularly useful after long occlusions or scene transitions.

\subsubsection{Gated Fusion of ST/LT Responses}
The final memory response $Z_{t,o}$ is computed
by a gated linear combination of the two responses:
\begin{equation}
\begin{aligned}
Z_{t,o} = \alpha Z^{ST}_{t,o} + (1-\alpha) Z^{LT}_{t,o},
\qquad 0 \le \alpha \le 1.
\end{aligned}
\label{eq:gated_fusion}
\end{equation}
Here, $\alpha$ controls the contribution of short-term evidence.
A larger $\alpha$ emphasizes recent appearance cues and quick adaptation,
while a smaller $\alpha$ increases reliance on long-range identity context.
The fused response $Z_{t,o}$ is passed to the SAM2 decoder
to produce the mask prediction $\hat{y}_{t,o}$.

\subsection{Text-Aware Memory (TAM)}
\label{sec:tam}
Dual Memory stabilizes temporal evidence structurally,
but purely visual memory can still be unreliable
under prolonged occlusion,
severe deformation,
or insufficient initial appearance cues.
TAM addresses this by introducing a compact \emph{word-level} semantic anchor
and using it to modulate memory usage based on semantic compatibility.

\subsubsection{TAM Word Selection Policy}
\label{sec:tam_word_policy}
We extract a representative word for each object $o$ using only the first $N$ frames (here $N=5$).
For each frame $t \in \{1,\ldots,N\}$, we crop the target region using the bounding box of the available mask
(with a 2-pixel padding on each side), and feed the crop into BLIP v1~\cite{BLIP}.
To encourage noun-like, word-level predictions, we run BLIP in a prompt-conditioned generation mode with a fixed prefix
\texttt{``a photo of''}, and retrieve the top-$K$ candidate tokens for the next word along with their confidence scores
$\{(\tilde{w}^{(k)}_{t,o},\,c^{(k)}_{t,o})\}_{k=1}^{K}$ (here $K=50$).
We use the ground-truth mask at $t=1$ and predicted masks for $t \ge 2$ to define the crop.

Not all candidates are informative at the word level (e.g., articles or numerals).
We therefore apply a lightweight lexical filter based on a small set of non-informative tokens, denoted by $\mathcal{S}$
(e.g., \texttt{a}, \texttt{the}, \texttt{one}, \texttt{two}).
Specifically, we scan the top-$K$ candidates in descending rank order and select the first token that is not filtered out:
\begin{equation}
\begin{aligned}
k^* &= \min\Bigl\{\,k \;\Bigm|\; \tilde{w}^{(k)}_{t,o} \notin \mathcal{S}\Bigr\},\\
w_{t,o} &= \tilde{w}^{(k^*)}_{t,o}, \qquad
c_{t,o} = c^{(k^*)}_{t,o}.
\end{aligned}
\label{eq:frame_word}
\end{equation}

We then select the final word as the one obtained from the frame with the highest confidence:
\begin{equation}
\begin{aligned}
t^\ast &= \mathop{\text{argmax}}_{t \in \{1,\ldots,N\}} c_{t,o},\\
w_o^\ast &= w_{t^\ast,o}.
\end{aligned}
\label{eq:word_select}
\end{equation}
The resulting cue $w_o^\ast$ captures a coarse semantic concept (e.g., \texttt{person}, \texttt{car})
beyond fine-grained instance appearance, and is fixed after frame $N$.

\subsubsection{TAM Gate}
\label{sec:tam_gate}
\paragraph{Text embedding.}
We obtain a Matryoshka text embedding $\tilde{t}_o \in \mathbb{R}^{d_{\text{clip}}}$ from JINA-CLIP~\cite{JINA-CLIP} using the prompt template
\texttt{``\{word\}''}.
Let $d_k$ denote the channel dimension of SAM2 memory keys.
To match dimensions without training, we truncate the text embedding to the first $d_k$ channels:
$t_o = \mathrm{Trunc}_{d_k}(\tilde{t}_o)$,
then apply $\ell_2$ normalization, $\bar{t}_o = t_o / \|t_o\|_2$, before computing cosine similarity with memory keys.

\paragraph{Semantic similarity to memory tokens.}
Each memory slot $m_{t,o}$ consists of a set of key/value tokens produced by the SAM2 memory encoder.
We flatten tokens across stored frames and spatial locations, and index them by
$i \in \{1,\dots,N_{\mathrm{tok}}\}$.
For each memory token with key vector $k_{o,i}$,
we compute cosine similarity using the $\ell_2$-normalized vectors $\bar{k}_{o,i} = k_{o,i}/\|k_{o,i}\|_2$ and $\bar{t}_o$ (the truncated and normalized text embedding from above):
\begin{equation}
\begin{aligned}
s_{o,i}
= \bar{k}_{o,i}^{\top}\bar{t}_o
= \frac{k_{o,i}^{\top} t_o}{\| k_{o,i}\|_2 \, \| t_o\|_2}.
\end{aligned}
\label{eq:cos_sim}
\end{equation}
This measures how semantically compatible each memory token is with the target word concept.

\paragraph{Text-gated reweighting of memory.}
We convert the similarity score into a non-negative scalar gate
using ReLU and a fixed modulation strength $\lambda_{\mathrm{word}}$.
We add a constant offset to make the modulation conservative in a residual manner: the gate is always lower-bounded by $1$,
so TAM never attenuates memory tokens and only upweights semantically compatible ones, reverting to the baseline behavior when the text cue is uninformative (i.e., $s_{o,i}\le 0$).
\begin{equation}
\begin{aligned}
g_{o,i}
= 1 + \lambda_{\mathrm{word}} \cdot \mathrm{ReLU}(s_{o,i}),
\qquad \lambda_{\mathrm{word}} = 0.1.
\end{aligned}
\label{eq:text_gate}
\end{equation}
The constant term $1$ ensures that when the text signal is absent
(or $\mathrm{ReLU}(s_{o,i})=0$),
the original memory contribution is preserved.
We then reweight the memory key (and in practice, the value as well)
using the same gate:
\begin{equation}
\begin{aligned}
k^\ast_{o,i} = g_{o,i} \, k_{o,i},
\qquad
v^\ast_{o,i} = g_{o,i} \, v_{o,i}.
\end{aligned}
\label{eq:kv_reweight}
\end{equation}

\paragraph{Where TAM is applied.}
We apply TAM to \emph{memory tokens} before attention by reweighting key/value tokens in both ST and LT banks.
TAM is enabled only after the word cue is fixed, i.e., from frame $t=N+1$ onward (the 6th frame when $N=5$):
we re-compute $Z^{ST}_{t,o}$ and $Z^{LT}_{t,o}$ using the reweighted tokens,
and then fuse them by Eq.~(\ref{eq:gated_fusion}).
Finally, the fused response is decoded to obtain $\hat{y}_{t,o}$, after which we update the ST/LT memory banks for the next step.
The full streaming inference procedure is summarized in Algorithm~\ref{alg:sam2dual} in the Appendix.


\subsection{Hyperparameters (Reference Setting)}
\label{sec:hyperparams}
Unless otherwise stated, we use the reference configuration in Table~\ref{tab:hyperparams}.
We resize each frame so that its longer side is $1024$ and initialize tracking with a first-frame mask prompt.
For Dual Memory, the short-term (ST) bank retains the most recent $K{=}7$ slots, while the long-term (LT) bank
stores up to $L{=}7$ slots and is updated every $m{=}10$ frames.
The ST/LT readouts are fused with coefficient $\alpha{=}0.63$ (Eq.~(\ref{eq:gated_fusion})).
For TAM, we extract a fixed word cue from the first $N{=}5$ frames using the top-$K_{\text{word}}{=}50$
BLIP candidates per frame with a lightweight lexical filter (non-informative token set $\mathcal{S}$),
and apply text modulation with $\lambda_{\mathrm{word}}{=}0.1$ to both memory keys and values
(Eqs.~(\ref{eq:text_gate})--(\ref{eq:kv_reweight})).

\begin{table}[h]
  \centering
  \caption{Reference setting used for all experiments.}
  \label{tab:hyperparams}
  \small
  \setlength{\tabcolsep}{6pt}      
  \renewcommand{\arraystretch}{1.05} 
  \begin{tabular}{ll}
  \toprule
  \textbf{Component} & \textbf{Setting} \\
  \midrule
  Input & long side $1024$, mask@t1 \\
  Dual Memory & $K{=}7$, $L{=}7$, $m{=}10$, $\alpha{=}0.63$ \\
  TAM & $N{=}5$, $K_{\text{word}}{=}50$, $\lambda_{\mathrm{word}}{=}0.1$, gate K\&V \\
  \bottomrule
  \end{tabular}
  \end{table}

\noindent\textbf{Compute.}
All experiments were conducted on a single NVIDIA Tesla V100 GPU with 40\,GB VRAM.
We use Python 3.12, PyTorch 2.6, and CUDA 12.8.
We use the official SAM2 pretrained checkpoint \texttt{sam2\_hiera\_large.pt} and do not perform any additional training or finetuning.
We report OOM (out-of-memory) when a method exceeds the available GPU memory under this reference setting.


\section{Results}

\subsection{Datasets}
\label{sec:datasets}
All experiments are conducted in a \emph{training-free} manner: we perform streaming inference only and never update model weights.
Table~\ref{tab:datasets_usage} summarizes the datasets/splits and their roles in our study.

\paragraph{Main results vs.\ Appendix results.}
We tune hyperparameters on DAVIS\_train and LVOSv2\_train, and report \textbf{main results} on MOSEv2, LVOSv2\_val, and PUMaVOS\_val following the official protocols.
We additionally report \textbf{Appendix results} on several splits; however, some of these were obtained \emph{before} finalizing the reference hyperparameters.
This is because the corresponding evaluation server became unavailable after the hyperparameter sweep, preventing re-evaluation with the final setting.
We therefore treat these numbers as supplementary and report them separately in the Appendix.

\begin{table}[h]
\centering
\caption{Datasets/splits and their roles in our study (all inference-only).}
\label{tab:datasets_usage}
\small
\setlength{\tabcolsep}{6pt}
\renewcommand{\arraystretch}{1.08}
\begin{tabular}{lll}
\toprule
\textbf{Dataset / split} & \textbf{Role} & \textbf{Used in} \\
\midrule
DAVIS\_train~\cite{DAVIS_Ori,Davis} & HP tuning & tuning only \\
DAVIS\_dev~\cite{DAVIS_Ori,Davis} & Appendix eval & Appendix only \\
YouTube-VOS\_val~\cite{YouTube-VOS} & Appendix eval & Appendix only \\
MOSEv2~\cite{MOSEv2} & evaluation & main \& Appendix \\
LVOSv1\_val~\cite{LVOSv1} & Appendix eval & Appendix only \\
LVOSv2\_train~\cite{LVOSv2} & HP tuning & tuning only \\
LVOSv2\_val~\cite{LVOSv2} & evaluation & main \& Appendix \\
PUMaVOS\_val~\cite{xmem++} & evaluation & main \& Appendix \\
\bottomrule
\end{tabular}
\end{table}

\begin{table}[h]
\centering
\caption{Summary statistics of the \emph{main evaluation} benchmarks used for the main results.}
\label{tab:main_dataset_stats}
\small
\setlength{\tabcolsep}{5pt}
\renewcommand{\arraystretch}{1.08}
\begin{tabular}{lccc}
\toprule
\textbf{Statistic} & \textbf{MOSEv2} & \textbf{LVOSv2\_val} & \textbf{PUMaVOS\_val} \\
\midrule
Year & 2025 & 2024 & 2023 \\
Avg. frames/video & 154 & 472 & 883 \\
Videos & 433 & 140 & 24 \\
Objects & 575 & 239 & 26 \\
\bottomrule
\end{tabular}
\end{table}

\paragraph{Dataset characteristics.}
\textbf{MOSEv2}~\cite{MOSEv2} consists of complex real-world videos with multiple objects, cluttered backgrounds, and frequent occlusions, where ambiguous visual evidence can lead to identity confusion.
\textbf{LVOSv2}~\cite{LVOSv2} targets long-term VOS with long sequences, large temporal gaps, and re-appearance events, making it suitable for assessing long-horizon drift and recovery.
\textbf{PUMaVOS}~\cite{xmem++} provides fine-grained and challenging annotations (e.g., detailed structures/parts), serving as an additional stress test for long-video robustness under identity ambiguity.
\textbf{DAVIS}~\cite{DAVIS_Ori,Davis} is a high-quality short-term benchmark with pixel-accurate annotations, commonly used for hyperparameter tuning under appearance changes, fast motion, and partial occlusions.
\textbf{YouTube-VOS}~\cite{YouTube-VOS} is a large-scale benchmark with diverse categories and videos; we report supplementary results in the Appendix.
\textbf{LVOSv1}~\cite{LVOSv1} is an earlier long-term VOS benchmark focusing on large temporal gaps and re-appearance events, also reported in the Appendix.

For all datasets, we follow the official evaluation protocols and report results on the standard splits provided by each benchmark.

\subsection{Evaluation Metrics}
\label{sec:metrics}
Following common VOS practice~\cite{DAVIS_Ori}, we report the combined \textbf{J\&F} score,
which averages region similarity (\textbf{J}) and boundary accuracy (\textbf{F}).

\paragraph{Region similarity (J).}
For each frame, J is defined as the intersection-over-union (IoU) between the predicted mask
$\hat{y}$ and the ground-truth mask $y$:
\begin{equation}
\begin{aligned}
\mathcal{J}(\hat{y}, y) = \frac{|\hat{y} \cap y|}{|\hat{y} \cup y|}.
\end{aligned}
\label{eq:j_metric}
\end{equation}

\paragraph{Boundary accuracy (F).}
F measures contour quality using the F-measure between predicted and ground-truth boundaries.
Specifically, boundaries are extracted from masks and matched within a small tolerance,
then precision and recall are computed to obtain:
\begin{equation}
\begin{aligned}
\mathcal{F} =
\frac{2 \cdot \mathrm{Precision} \cdot \mathrm{Recall}}
{\mathrm{Precision} + \mathrm{Recall}}.
\end{aligned}
\label{eq:f-metric}
\end{equation}

\paragraph{Combined score (J\&F).}
We compute the per-frame score $\frac{1}{2}(\mathcal{J}+\mathcal{F})$ and average across frames.
For multi-object sequences, scores are additionally averaged across objects:
\begin{equation}
\begin{aligned}
\mathrm{J\&F} = \frac{1}{2}\left(\overline{\mathcal{J}}+\overline{\mathcal{F}}\right),
\end{aligned}
\label{eq:jf-metric}
\end{equation}
where $\overline{\cdot}$ denotes averaging over frames (and objects when applicable)~\cite{DAVIS_Ori}.

\subsection{Quantitative results on long-term benchmarks}
Table~\ref{tab:main_results_jf_long} reports the quantitative comparison (J\&F) on three long-term VOS benchmarks
(MOSEv2, LVOSv2, and PUMaVOS).
Conventional memory-propagation methods struggle to scale to long horizons:
DeAOT runs out of memory (OOM) on all three datasets, while XMem++ and Cutie show substantial degradation~\cite{DEAOT,xmem++,Cutie,SAMURAI,Him2sam},
highlighting the practical difficulty of long-video inference under limited memory budgets.
With the emergence of SAM2, the evaluation focus in promptable and foundation-style VOS has largely shifted toward SAM2-centered baselines, since earlier memory-propagation approaches are less practical under long-video settings.

Among promptable foundation-style baselines, SAM2 provides a strong zero-shot reference
(49.33 on MOSEv2, 82.1 on LVOSv2, and 81.0 on PUMaVOS).
Our full model, \textbf{SAM2Dual}, achieves the best overall stability on long videos,
improving MOSEv2 from 49.33 to 50.65 and LVOSv2 from 82.1 to 83.1.
On PUMaVOS, SAM2Dual remains competitive at 80.6, which is slightly below SAM2 (81.0)
but still above SAMURAI and HiM2SAM.
Notably, SAMURAI and HiM2SAM are primarily tracking-oriented variants, and although they improve temporal tracking robustness, they do not show equally strong segmentation quality on these long-term VOS benchmarks.
These results indicate that restructuring streaming memory at inference time is effective
for mitigating long-horizon drift while preserving strong zero-shot performance.

We attribute the gain of SAM2Dual primarily to its ability to prevent recent, unreliable
predictions from saturating the memory state, thereby reducing progressive drift under
long occlusions and appearance shifts.
The semantic cue from TAM provides an additional constraint that is especially helpful
when visual evidence is weak or confusable, improving identity consistency via semantic compatibility.

\begin{table}[!t]
    \centering
    \caption{J\&F (\%) on long-term VOS benchmarks (MOSEv2, LVOSv2, and PUMaVOS; higher is better). Best in \textbf{bold} and second best \underline{underlined}; OOM indicates out-of-memory.}
    \label{tab:main_results_jf_long}
    \setlength{\tabcolsep}{8pt}
    \renewcommand{\arraystretch}{1.30}
    \begin{tabular}{lccc}
    \hline
    \textbf{Method} & \textbf{MOSEv2} & \textbf{LVOSv2} & \textbf{PUMaVOS} \\
    \hline
    DeAOT \cite{DEAOT}                 & OOM   & OOM  & OOM  \\
    XMem++ \cite{xmem++}                & 36.20 & 62.6 & 62.3 \\
    Cutie \cite{Cutie}                 & 41.90 & 77.7 & 65.9 \\
    \hline
    SAM2 \cite{Sam2}                  & \underline{49.33} & \underline{82.1} & \textbf{81.0} \\
    SAMURAI \cite{SAMURAI}               & 47.82 & \underline{82.1} & 79.4 \\
    HiM2SAM \cite{Him2sam}               & 45.90 & 81.7 & 78.5 \\
    \hline
    SAM2Dual(ours)              & \textbf{50.65} & \textbf{83.1} & \underline{80.6} \\
    \hline
    \end{tabular}
\end{table}

\subsection{Effect on long sequences (MOSEv2)}

\begin{figure}[h]
  \centering
  \includegraphics[width=\linewidth]{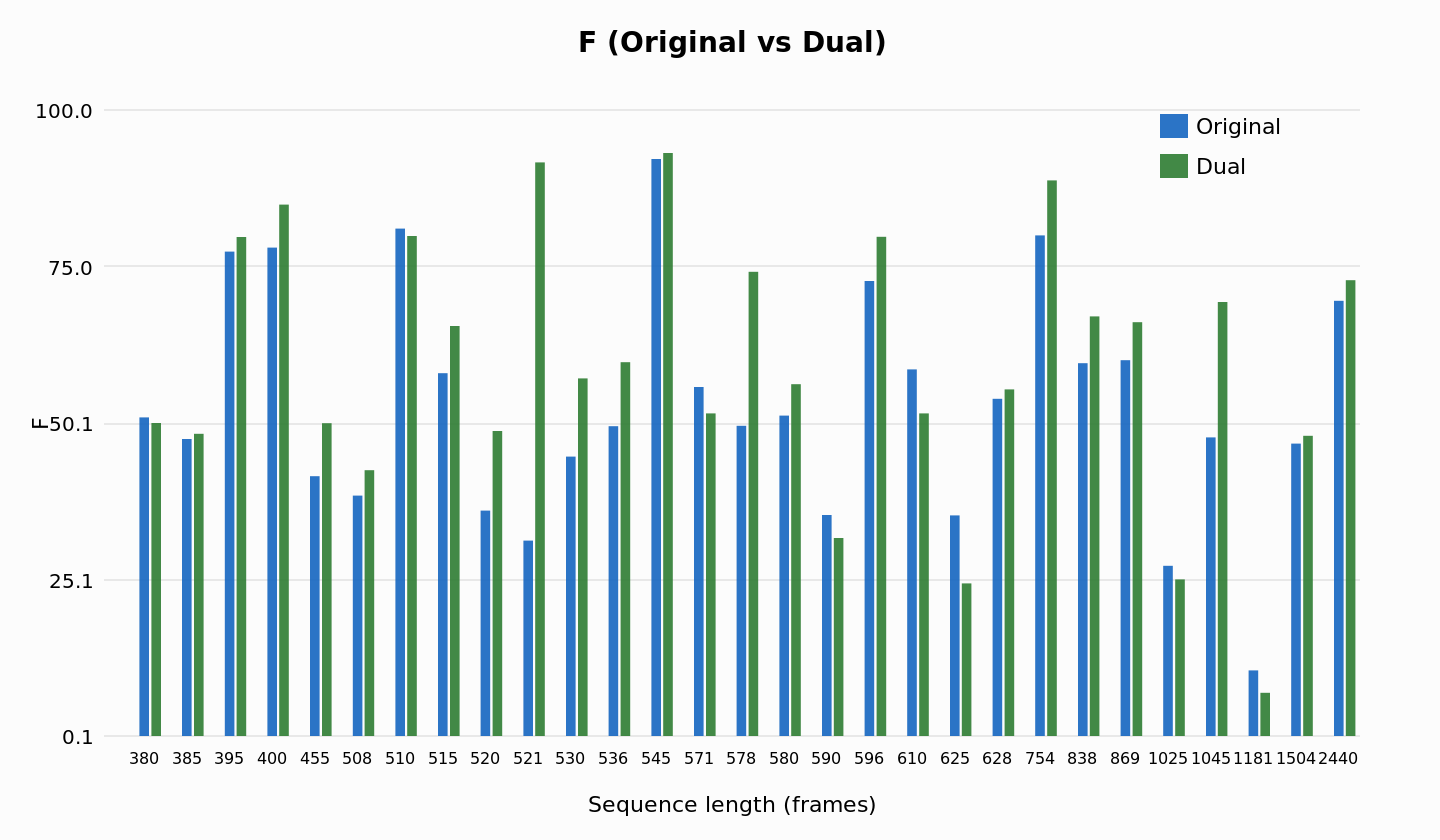}
  \caption{Per-sequence $F$ scores ordered by sequence length.}
  \label{fig:long_F_changed}
\end{figure}

\begin{figure}[h]
  \centering
  \includegraphics[width=\linewidth]{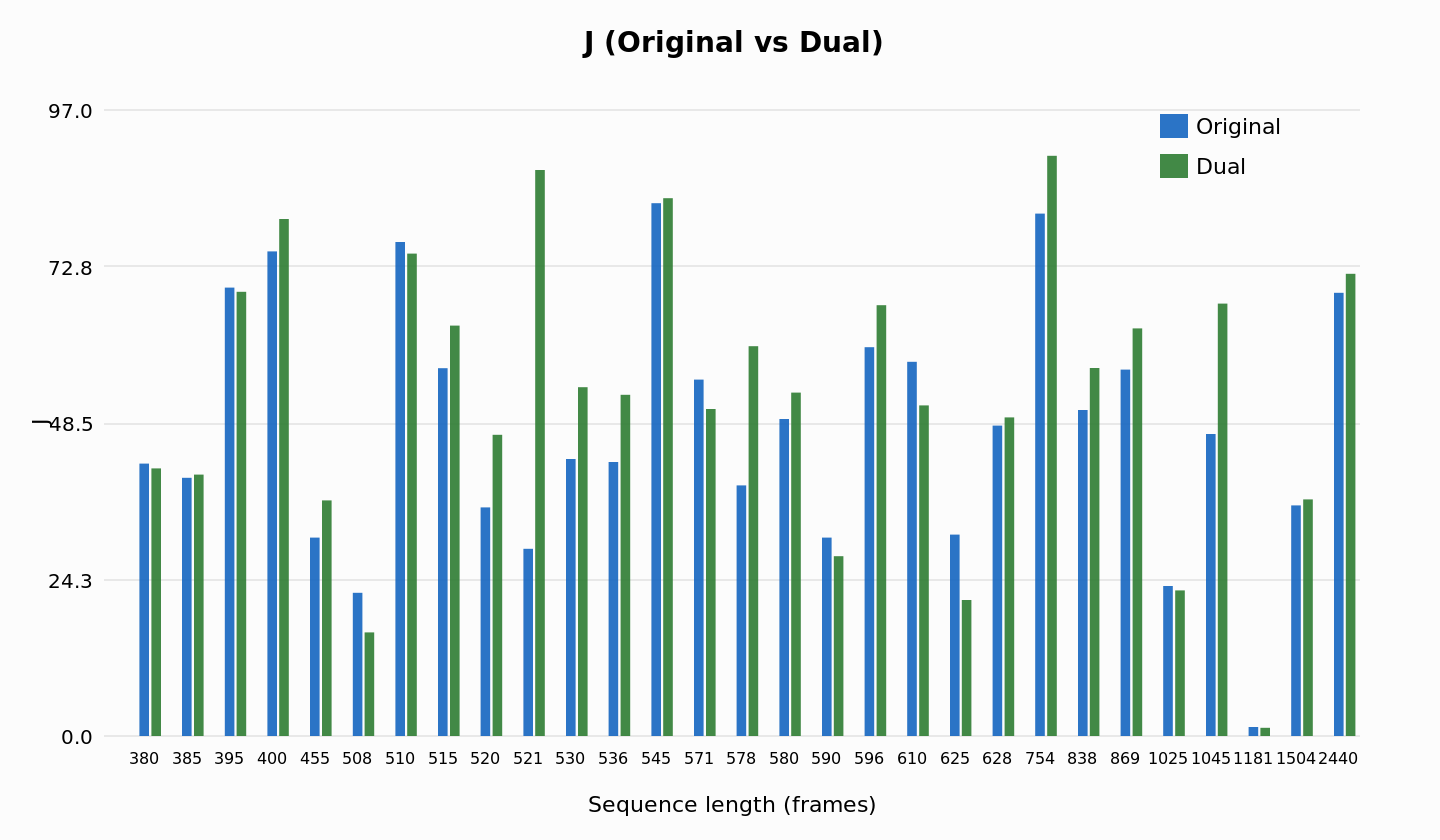}
  \caption{Per-sequence $J$ scores ordered by sequence length.}
  \label{fig:long_J_changed}
\end{figure}

We compare SAM2 and SAM2Dual on the top 10\% longest sequences of MOSEv2 (Fig.~3 and Fig.~4).
Among the sequences where score differences are observed, SAM2Dual attains higher $F$ and $J$ scores than the original SAM2 for \emph{most} sequences.
In several cases, the gains are substantial, while a smaller number of sequences exhibit degradations, indicating sequence-dependent behavior on long videos.
Overall, these results suggest that the proposed Dual Memory design can help preserve object representations over extended temporal horizons.

\subsection{Dual Memory vs.\ TAM}

\begin{table}[!h]
  \centering
  \caption{J\&F (\%) on MOSEv2, LVOSv2\_val, and PUMaVOS\_val. Best in \textbf{bold}, second best \underline{underlined}.}
  \label{tab:ablation_dual_tam_transposed}
  \setlength{\tabcolsep}{6pt}
  \renewcommand{\arraystretch}{1.15}
  \begin{tabular}{lccc}
  \hline
  \textbf{Model / Setting} & \textbf{MOSEv2} & \textbf{LVOSv2\_val} & \textbf{PUMaVOS\_val} \\
  \hline
  \textbf{Dual only}              & 50.4  & 82.9 & 80.7 \\
  \textbf{TAM only}               & 49.33 & 82.3 & 81.0 \\
  \textbf{Dual + TAM}             & 50.65 & 83.1 & 80.6 \\
  \hline
  \end{tabular}
\end{table}

Table~\ref{tab:ablation_dual_tam_transposed} analyzes the complementary roles of Dual Memory and TAM.
Dual-only provides the main improvement on MOSEv2 (50.4) and LVOSv2\_val (82.9),
confirming that temporal structuring is the primary driver of long-horizon stability.
TAM-only remains close to the SAM2 baseline on MOSEv2 (49.33) and yields a modest improvement on LVOSv2\_val (82.3).
On PUMaVOS\_val, TAM-only reaches 81.0, indicating that semantic identity cues can be helpful in this setting,
although the gain is limited and does not clearly carry over to the full model.
Combining both modules produces the best results on MOSEv2 and LVOSv2\_val, reaching 50.65 and 83.1, respectively.
These trends indicate that Dual Memory mainly improves long-horizon stability through multi-scale temporal evidence,
while TAM plays a complementary role by supporting identity-consistent retrieval when visual evidence is weak or ambiguous.

\begin{figure*}[!t]
  \centering
  \includegraphics[width=\textwidth]{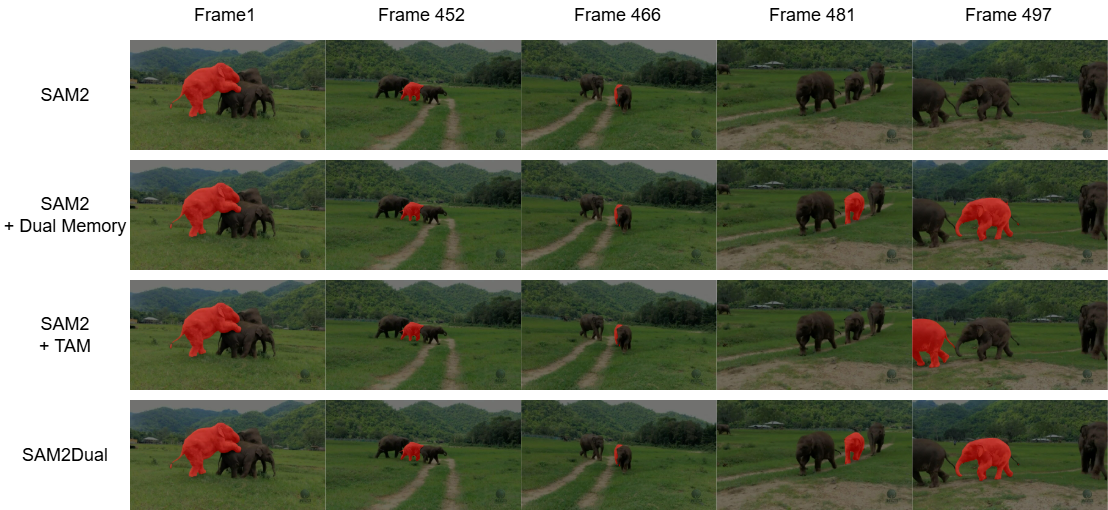}
  \caption{Qualitative ablation on a long elephant sequence.
  Rows (top to bottom): SAM2, SAM2 + Dual Memory, SAM2 + TAM, and SAM2Dual.
  SAM2 progressively erodes the mask and loses the target in later frames,
  whereas SAM2Dual maintains stable target coverage over time.}
  \label{fig:results_el}
\end{figure*}

\begin{figure*}[!t]
  \centering
  \includegraphics[width=\textwidth]{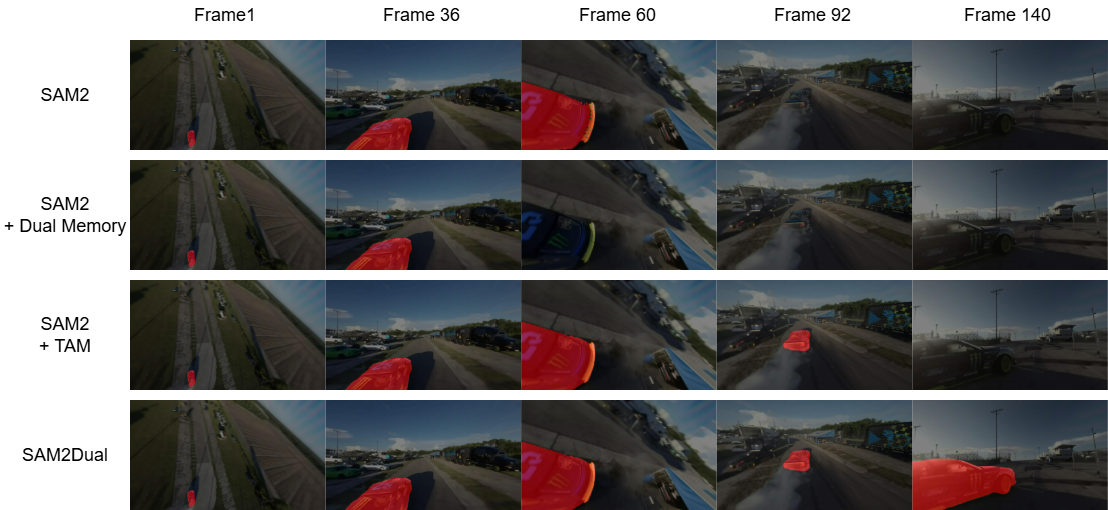}
  \caption{Qualitative ablation on a long car sequence.
  Rows (top to bottom): SAM2, SAM2 + Dual Memory, SAM2 + TAM, and SAM2Dual.
  SAM2 progressively erodes the mask and loses the target in later frames,
  whereas SAM2Dual maintains stable target coverage over time.}
  \label{fig:results_car}
\end{figure*}

Figs.~\ref{fig:results_el} and~\ref{fig:results_car} provide qualitative comparisons on long videos.
In each figure, rows (top to bottom) correspond to \textbf{SAM2}, \textbf{SAM2 + Dual Memory}, \textbf{SAM2 + TAM}, and \textbf{SAM2Dual},
and columns show selected frames over time.
Overall, SAM2 tends to exhibit progressive drift or target loss in later frames,
while SAM2Dual preserves more stable masks by combining multi-scale temporal evidence (Dual Memory)
with text-aware reweighting (TAM), consistent with the complementary trends in Table~\ref{tab:ablation_dual_tam_transposed}.

In Fig.~\ref{fig:results_el}, we compare SAM2 and our variants on a long sequence containing a herd of elephants.
The video involves large scale changes, frequent inter-instance occlusions, and strong perspective shifts in later frames, making identity preservation difficult.
While SAM2 produces an accurate mask in early frames (Frame~1), it gradually shrinks and deforms the mask around mid-sequence (Frames~452--466) and often fails to recover the target after re-appearance, sometimes drifting to background or other instances.
Dual Memory mitigates this drift by retaining long-term appearance cues across occlusions.
TAM-only, however, is limited by category-level semantics (\texttt{elephant}) and may suffer identity switches when multiple similar instances are present.
SAM2Dual combines both modules and yields the most consistent masks throughout the sequence.

In Fig.~\ref{fig:results_car}, the racing-track sequence exhibits repeated dust occlusions caused by a preceding vehicle, yielding very limited visual cues for the target car in early frames.
The SAM2 baseline becomes unstable when dust occlusion occurs before sufficient visual evidence has been accumulated, resulting in unreliable segmentation in the mid sequence.
Dual-only can also fail in this setting: when the initial appearance evidence is weak, incorrect or incomplete cues can be written into both ST and LT memories and subsequently reinforced, eventually causing the tracker to lose the target.
By contrast, TAM-only benefits from the generic semantic cue (\texttt{car}), allowing it to follow the vehicle more stably in the early stage despite weak visual evidence; however, without explicit temporal structuring of memory, its predictions can again become unstable in later frames.
SAM2Dual alleviates both issues: TAM reduces early-stage errors by providing semantic anchoring under ambiguity, and Dual Memory then preserves reliable appearance evidence over time, enabling the most consistent tracking of the target car up to the final frame despite repeated dust occlusions.

\subsection{Hyperparameter sensitivity and reference setting}

\begin{table}[h]
  \centering
  \caption{Hyperparameter sweep on \textbf{LVOSv2 train} (J\&F, \%).
We vary $\lambda_{\mathrm{word}}\in\{0.10,0.05,0.01\}$, $\alpha\in\{0.50,0.63,0.75\}$, and $m\in\{5,10,15\}$.
Best is in \textbf{bold}.}
  \label{tab:lvos_full_sweep}
  \small
  \setlength{\tabcolsep}{5pt}
  \renewcommand{\arraystretch}{1.05}
  \begin{tabular*}{\linewidth}{@{\extracolsep{\fill}} c c ccc}
  \toprule
  $\lambda_{\mathrm{word}}$ & $\alpha$ & $m{=}5$ & $m{=}10$ & $m{=}15$ \\
  \midrule
  \multirow{3}{*}{0.10}
  & 0.50 & 92.20 & 92.30 & 92.10 \\
  & 0.63 & 92.50 & \textbf{92.80} & 92.40 \\
  & 0.75 & 92.20 & 92.20 & 92.00 \\
  \midrule
  \multirow{3}{*}{0.05}
  & 0.50 & 92.40 & 92.30 & 92.50 \\
  & 0.63 & 92.00 & 91.60 & 92.40 \\
  & 0.75 & 92.20 & 91.70 & 92.20 \\
  \midrule
  \multirow{3}{*}{0.01}
  & 0.50 & 92.00 & 92.00 & 92.30 \\
  & 0.63 & 92.00 & 91.50 & 92.10 \\
  & 0.75 & 91.90 & 91.70 & 91.60 \\
  \bottomrule
  \end{tabular*}
  
  \vspace{2pt}
  \scriptsize{\textbf{Baseline (original)} on LVOSv2 train: J\&F = 92.3}
  \end{table}

\begin{table}[h]
  \centering
  \caption{Hyperparameter sweep on \textbf{DAVIS train} (J\&F, \%).
We vary $\lambda_{\mathrm{word}}\in\{0.10,0.05,0.01\}$, $\alpha\in\{0.50,0.63,0.75\}$, and $m\in\{5,10,15\}$.
Best is in \textbf{bold}.}
  \label{tab:davis_full_sweep}
  \small
  \setlength{\tabcolsep}{5pt}
  \renewcommand{\arraystretch}{1.05}
  \begin{tabular*}{\linewidth}{@{\extracolsep{\fill}} c c ccc}
  \toprule
  $\lambda_{\mathrm{word}}$ & $\alpha$ & $m{=}5$ & $m{=}10$ & $m{=}15$ \\
  \midrule
  \multirow{3}{*}{0.10}
  & 0.50 & 90.60 & 90.80 & 90.60 \\
  & 0.63 & \textbf{90.90} & \textbf{90.90} & 90.60 \\
  & 0.75 & \textbf{90.90} & 90.60 & 90.50 \\
  \midrule
  \multirow{3}{*}{0.05}
  & 0.50 & 90.60 & 90.80 & 90.60 \\
  & 0.63 & 90.60 & \textbf{90.90} & 90.50 \\
  & 0.75 & \textbf{90.90} & 90.60 & 90.50 \\
  \midrule
  \multirow{3}{*}{0.01}
  & 0.50 & 90.60 & 90.80 & 90.50 \\
  & 0.63 & 90.50 & 90.70 & 90.50 \\
  & 0.75 & 90.80 & 90.60 & 90.20 \\
  \bottomrule
  \end{tabular*}
  
  \vspace{2pt}
  \scriptsize{\textbf{Baseline (original)} on DAVIS train: J\&F = 90.5}
  \end{table}

  Tables~\ref{tab:lvos_full_sweep} and~\ref{tab:davis_full_sweep} summarize sensitivity to the Dual Memory update interval $m$, the fusion gate $\alpha$, and the TAM modulation strength $\lambda_{\mathrm{word}}$ on LVOSv2 train and DAVIS train, respectively.
  Performance varies smoothly across a broad range of settings, suggesting that SAM2Dual does not rely on fragile tuning.
  To ensure that the proposed method maintains performance not only on long videos but also on short videos, we select hyperparameters based on both DAVIS (short-term) and LVOSv2 (long-term) training splits.
  We adopt the best-performing reference setting shared across these datasets, namely $\alpha{=}0.63$, $\lambda_{\mathrm{word}}{=}0.1$, and $m{=}10$.

\subsection{Failure case on PUMaVOS.}

\begin{figure}[!h]
  \centering
  \includegraphics[width=\linewidth]{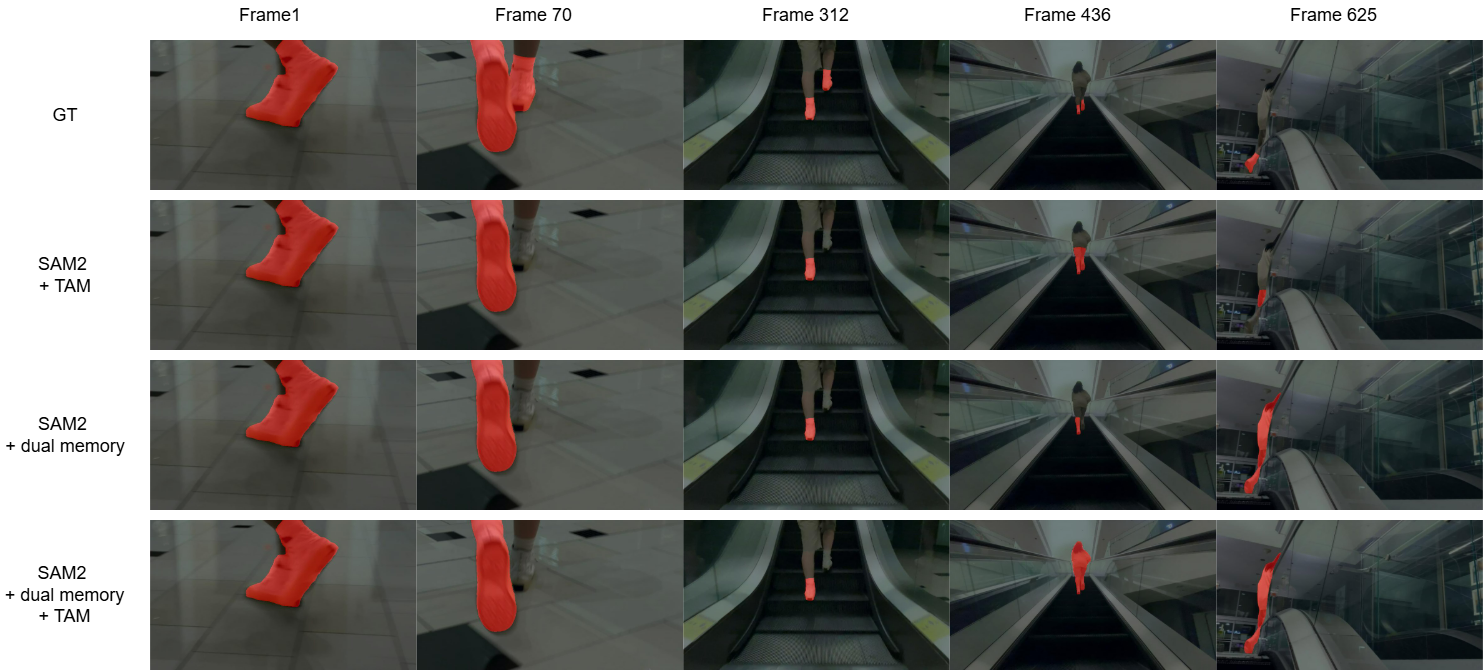}
  \caption{Representative failure case on PUMaVOS with a fine-grained, part-level target (shoe), where SAM2Dual drifts and switches identity to the full person in later frames.}
  \label{fig:results_fail}
\end{figure}

Fig.~\ref{fig:results_fail} shows a representative failure case on PUMaVOS with a fine-grained, part-level target (shoe).
While SAM2+TAM remains relatively localized around the target, both variants with Dual Memory gradually drift toward the full person in later frames (e.g., Frames~436 and~625).
Although Dual Memory improves overall long-horizon stability, this example suggests that the current fixed periodic memory-update policy can be less robust in certain part-level scenarios.
In such cases, an inaccurate intermediate prediction may be written into long-term memory and subsequently affect later retrieval, making precise recovery more difficult.
This tendency appears particularly relevant for part-level targets, where a small local region can be absorbed into a larger semantically related instance such as the whole person.
As a future direction, adaptive memory-update strategies---for example, confidence-aware, consistency-aware, or event-triggered updates instead of fixed periodic writes---may further improve robustness on challenging long-horizon cases.

\section{Conclusion}
In this paper, we presented \textbf{SAM2Dual}, a \textbf{training-free, plug-and-play} inference-time enhancement for SAM2 that improves robustness in \textbf{long-term video object segmentation} without updating model weights.
SAM2Dual mitigates long-horizon drift and identity instability by restructuring streaming memory across temporal scales and injecting lightweight semantic cues into memory retrieval.

We introduced a \textbf{Dual Memory} design that separates (i) a short-term memory for rapid local adaptation and (ii) an interval-sampled long-term memory updated every $m$ frames to preserve global identity cues, and combine them via a simple gated fusion.
We also proposed \textbf{Text-Aware Memory (TAM)}, which extracts a compact \emph{word-level} cue from early frames and uses text embeddings to reweight memory keys/values based on semantic compatibility.

Experiments show that SAM2Dual consistently improves long-video stability on challenging benchmarks, raising J\&F from 49.33 to 50.65 on MOSEv2 and from 82.1 to 83.1 on LVOSv2, while maintaining competitive performance on short-sequence datasets.
Overall, the results suggest that long-term failures in streaming segmentation are often driven by monolithic, recent-frame-dominated memory, and that \textbf{multi-scale memory structuring with semantic reweighting} provides an effective training-free remedy.


%



%

\appendices

\section{Training-Free Inference Pipeline (Algorithm)}
\label{app:inference_algorithm}

{\renewcommand{\figurename}{Algorithm}%
\setcounter{savedfigure}{\value{figure}}%
\setcounter{figure}{\value{algorithm}}%
\begin{figure}[!h]
\centering
\footnotesize
\begin{algorithmic}[1]
\STATE \textbf{Input:} Video frames $\{x_t\}_{t=1}^{T}$, first-frame prompt mask $\{y_{1,o}\}_{o=1}^{O}$; hyperparameters $N,\alpha,K_{\text{word}},K,L,m$.
\STATE \textbf{Output:} Predicted masks $\{\hat{y}_{t,o}\}_{t=2}^{T}$ for each object $o$.
\STATE

\STATE \textbf{(0) Initialize memory:}
\FOR{each object $o$}
\STATE \quad Create slot $m_{1,o}$ from $(x_1, y_{1,o})$; initialize $M^{ST}_{1,o} \leftarrow \{m_{1,o}\}$ and $M^{LT}_{1,o} \leftarrow \{m_{1,o}\}$.
\STATE \quad Initialize word-cue buffer $\mathcal{B}_o \leftarrow \emptyset$ and set \texttt{word\_ready} $\leftarrow$ \texttt{false}.
\ENDFOR

\FOR{$t = 2$ \textbf{to} $T$}
\FOR{each object $o$}

\STATE \textbf{(1) Read ST/LT (TAM off/on):}
\STATE \quad Compute query $q_{t,o}$ from $x_t$.
\IF{$t \ge N+1$ \textbf{and} \texttt{word\_ready}}
\STATE \quad Compute cosine similarity $s_{o,i}$ and gate $g_{o,i}$ by Eqs.~(\ref{eq:cos_sim})--(\ref{eq:text_gate}).
\STATE \quad Reweight key/value tokens by Eq.~(\ref{eq:kv_reweight}) for both ST and LT; read with reweighted tokens (TAM on).
\ELSE
\STATE \quad Read with visual-only tokens (TAM off).
\ENDIF
\STATE \quad $Z^{ST}_{t,o} \leftarrow \mathrm{Attn}(q_{t,o}, M^{ST}_{t-1,o})$; \quad
       $Z^{LT}_{t,o} \leftarrow \mathrm{Attn}(q_{t,o}, M^{LT}_{t-1,o})$.

\STATE \textbf{(2) Fuse:} $Z_{t,o} \leftarrow \alpha Z^{ST}_{t,o} + (1-\alpha)Z^{LT}_{t,o}$.

\STATE \textbf{(3) Decode:} $\hat{y}_{t,o} \leftarrow \mathrm{Decoder}(Z_{t,o}, x_t)$.

\STATE \textbf{(4) Collect word cue (only for early frames):}
\IF{$t \le N$}
\STATE \quad Crop by the bounding box of $\hat{y}_{t,o}$ (2-pixel padding).
\STATE \quad Obtain top-$K_{\text{word}}$ word candidates from BLIP v1~\cite{BLIP}.
\STATE \quad Apply \texttt{good\_word} filter (e.g., remove \texttt{a}, \texttt{the}, \texttt{one}, \texttt{two}).
\STATE \quad Set $w_{t,o}$ as the first candidate that passes the filter (Eq.~(\ref{eq:frame_word})).
\STATE \quad Store $(w_{t,o}, c_{t,o})$ into buffer $\mathcal{B}_o$.
\ENDIF
\IF{$t = N$}
\STATE \quad Select $w_o^\ast$ from $\mathcal{B}_o$ by Eq.~(\ref{eq:word_select}).
\STATE \quad Embed $w_o^\ast$ into $t_o$ via JINA-CLIP~\cite{JINA-CLIP} using the prompt ``\{word\}''.
\STATE \quad Set \texttt{word\_ready} $\leftarrow$ \texttt{true}.
\ENDIF

\STATE \textbf{(5) Update memory:}
\STATE \quad Create slot $m_{t,o}$ from $(x_t, \hat{y}_{t,o})$.
\STATE \quad $M^{ST}_{t,o} \leftarrow M^{ST}_{t-1,o} \cup \{m_{t,o}\}$; keep the last $K$ slots.
\IF{$t \bmod m = 0$}
\STATE \quad $M^{LT}_{t,o} \leftarrow M^{LT}_{t-1,o} \cup \{m_{t,o}\}$; keep at most $L$ slots.
\ELSE
\STATE \quad $M^{LT}_{t,o} \leftarrow M^{LT}_{t-1,o}$.
\ENDIF

\ENDFOR
\ENDFOR
\end{algorithmic}
\caption{SAM2Dual inference pipeline (training-free).}
\label{alg:sam2dual}
\end{figure}
\setcounter{algorithm}{\value{figure}}%
\setcounter{figure}{\value{savedfigure}}%
}

\section{Additional Benchmark Results under a Previous Setting}
\label{prev_setting_results}

Our main tables report results under a single reference configuration for fair comparison.
For some benchmarks, evaluation under the current reference setting was not feasible due to
dataset/benchmark constraints in our environment.
For completeness, we provide the results obtained with a previous configuration.
These numbers are reported for reference and are not directly comparable to the main tables.

\begin{table}[!h]
\caption{J\&F (\%) comparison on VOS benchmarks under a previous setting.}
\label{tab:vos_results_prev}
\centering
\begin{threeparttable}
\scriptsize
\setlength{\tabcolsep}{3pt}
\renewcommand{\arraystretch}{1.05}
\begin{tabular}{lcccccc}
\toprule
\textbf{Method} & \textbf{DAVIS} & \textbf{YouTube-VOS} & \textbf{MOSEv2} & \textbf{LVOSv1} & \textbf{LVOSv2} & \textbf{PUMaVOS} \\
\midrule
DeAOT~\cite{DEAOT}        & 82.19 & 86.25 & OOM   & OOM   & OOM  & OOM  \\
XMem++~\cite{xmem++}      & 80.94 & 84.00 & 36.20 & 44.13 & 62.6 & 62.3 \\
Cutie~\cite{Cutie}        & 86.37 & 86.15 & 41.90 & 66.65 & 77.7 & 65.9 \\
\midrule
SAM2~\cite{Sam2}          & 88.59 & \underline{88.81} & 49.33 & 76.72 & 82.1 & \underline{81.0} \\
SAMURAI~\cite{SAMURAI}    & 88.98 & 88.51             & 47.82 & 75.90 & 82.1 & 79.4 \\
HiM2SAM~\cite{Him2sam}    & 88.77 & 88.64             & 45.90 & 67.31 & 81.7 & 78.5 \\
\midrule
TAM                       & 88.80 & \textbf{88.82}     & 49.39 & 76.33 & 82.1 & \textbf{81.1} \\
Dual Memory               & \underline{89.34} & 88.50 & \underline{50.40} & \underline{78.94} & \underline{82.9} & 80.7 \\
SAM2Dual                  & \textbf{89.40} & 88.56 & \textbf{50.57} & \textbf{80.41} & \textbf{83.0} & 80.8 \\
\bottomrule
\end{tabular}
\begin{tablenotes}
\scriptsize
\item \textbf{Bold}: best; \underline{Underline}: second best. OOM indicates out-of-memory.
\end{tablenotes}
\end{threeparttable}
\end{table}





\ifCLASSOPTIONcaptionsoff
  \newpage
\fi

%








\end{document}